\documentclass[10pt,twocolumn,letterpaper]{article}

\usepackage{cvpr}
\usepackage{times}
\usepackage{epsfig}
\usepackage{graphicx}
\usepackage{amsmath}
\usepackage{amssymb}
\usepackage{authblk}

\usepackage[mathcal]{euscript}

\def \argmax  {\arg\!\max}

\def\R{\mathbb{R}}

\def\soft{\mathsf{soft}}
\def\softnorm{\mathsf{softnorm}}

\def\hard{\mathsf{hard}}
\def\hardnorm{\mathsf{hardnorm}}
\def\LUTC{\mathsf{LUTC}}
\def\LUTSym{\mathsf{LUTSym}}
\def\LUTASym{\mathsf{LUTASym}}

\usepackage[pagebackref=true,breaklinks=true,letterpaper=true,colorlinks,bookmarks=false]{hyperref}

\cvprfinalcopy 

\def\cvprPaperID{2506} 

\ifcvprfinal\fi

\begin{document}

\title{End-to-End Supervised Product Quantization for Image Search and Retrieval}

\author[1]{Benjamin Klein}
\author[1,2]{Lior Wolf}
\affil[1]{The Blavatnik School of Computer Science, Tel Aviv University, Israel}
\affil[2]{Facebook AI Research}
\renewcommand\Authands{ and }

\maketitle

{\color{black}
\begin{abstract}
Product Quantization, a dictionary based hashing method, is one of the leading unsupervised hashing techniques. While it ignores the labels, it harnesses the features to construct look up tables that can approximate the feature space. In recent years, several works have achieved state of the art results on hashing benchmarks by learning binary representations in a supervised manner. This work presents Deep Product Quantization (DPQ), a technique that leads to more accurate retrieval and classification than the latest state of the art methods, while having similar computational complexity and memory footprint as the Product Quantization method. To our knowledge, this is the first work to introduce a dictionary-based representation that is inspired by Product Quantization and which is learned end-to-end, and thus benefits from the supervised signal. DPQ explicitly learns soft and hard representations to enable an efficient and accurate asymmetric search, by using a straight-through estimator. Our method obtains state of the art results on an extensive array of retrieval and classification experiments. 
\end{abstract}
\vspace{-4mm}

\section{Introduction}

Computer vision practitioners have adopted the Product Quantization (PQ) method~\cite{jegou2011product} as a leading approach for conducting Approximated Nearest Neighbor (ANN) search in large scale databases. However, the research community has recently shifted its focus toward using methods that compute Hamming distances on binary representations learned by supervised dictionary-free methods, and showed its superiority on the standard PQ techniques~\cite{subic}. In this work, we present a technique inspired by PQ and named Deep Product Quantization (DPQ) that outperforms previous methods on many known benchmarks. While standard PQ is learned in an unsupervised manner, our DPQ is learned in an end-to-end fashion, and benefits from the task-related supervised signal.

PQ methods decompose the embedding manifold into a Cartesian product of $M$ disjoint partitions, and quantize each partition into $K$ clusters. An input vector $x \in \R^{MD}$ is split into $M$ sub-vectors in~$\R^D$, $x= \left[ x_1, x_2, \dots, x_M \right]$ and then encoded by PQ as $z_x \in \{0,1\}^{M \cdot \log_{2}(K)}$. Each group of $\log_{2}(K)$ bits decodes the index $k\in \{1\dots K\}$ of the cluster to which the sub-vector belongs (note that the clusters vary between the subspaces). A representative vector $c_{m,k} \in \R^D$ is associated with each cluster $k$ of each partition $m$. An approximation of the original vector, $\tilde{x}$, can be readily reconstructed from $z_x$, by concatenating the representative vectors of the matching clusters. The common practice for training a PQ is to run K-means in an unsupervised manner on each partition, and to use the centroid of each cluster as the representative vector.

The advantages of using the PQ technique are the reduction in memory footprint and the acceleration of search time. The decomposition of the embedding into a Cartesian product of $M$ sub-vectors is the key ingredient in the effectiveness of PQ in reducing the retrieval search time, since it allows to compute the approximated distance of a pair of vectors, $x$ and $y$, directly from their compressed representations, $z_x$ and $z_y$, using look-up tables.
PQ methods can also achieve better retrieval performance by using an asymmetric search, in which the distance is computed between the source vector, $x$, and the compressed vector, $z_y$, with the same amount of computation as the symmetric search. 

Another common technique for ANN search is transforming the embedding into a binary representation, without using a dictionary, and performing the comparison using Hamming distance. Several works have achieved state of the art results on retrieval benchmarks, by learning the binary representation as part as of the classification model optimization in an end-to-end fashion. The binary representation is thus trained using a supervised signal and, therefore, the distance between two binary representations reflects the end goal of the system. 

In this work, we present a new technique, Deep Product Quantization, which to our knowledge, is the first to learn a compressed representation inspired by PQ, which is learned end-to-end and, therefore, benefits from the supervised signal. Our contributions include: 
(i) an end-to-end PQ approach for ANN search that exploits high-dimensional Euclidean distances through the use of a dictionary, instead of the Hamming distance,
(ii) learning soft and hard representations as part of the training to facilitate symmetric and asymmetric search, 
(iii) using a straight-through estimator to overcome the non-differential argmax function, which is essential for our hard representation, 
(iv) a new loss function named \emph{joint central loss}, which is inspired by the center loss~\cite{wen2016discriminative} but also decreases the discrepancy between the soft and the hard representations, 
(v) a normalization technique which improves the results for cross-domain category retrieval and (vi) a very extensive array of state of the art retrieval and classification results, to establish our claims using more literature protocols than any existing work.

\section{Related work}\label{sec:related}

Vector quantization techniques~\cite{gray1998quantization} have been used successfully in the past in many applications, including data compression, approximated nearest neighbor search, and clustering. The most classic technique is Vector Quantization (VQ) which, divides the space into $K$ clusters, by using an unsupervised clustering method, such as K-means. VQ allows encoding of each sample by $\log_{2}(K)$ bits, namely by encoding the identity of the cluster to which the sample belongs. By precomputing the euclidean distance between every two clusters and storing the results in a hash table with $O(K^2)$ entries, one can compute the approximated distance between every two samples in $O(1)$ time. 

Since the number of clusters grows exponentially as a function of the number of bits, one may expect the performance of VQ to improve as more bits are added. In practice, since VQ is learned using the K-means algorithm, a meaningful quantization of the space requires a number of samples, which is proportional to the number of clusters. Since the hash table grows quadratically in the number of clusters, it also becomes infeasible to use the hash table for large values of $K$. These reasons have limited the efficient usage of VQ to a small number of clusters. This limitation has an impact on the quantization error, i.e., the distance between the original vector and its matching centroid and, therefore, is a bottleneck in decreasing the quantization error and in improving the retrieval performance.

Product Quantization~\cite{jegou2011product} (PQ) is a clever technique to overcome the bottleneck of increasing the number of clusters with respect to VQ, while allowing an efficient computation of the approximated euclidean distance between two compressed representations, and reducing the quantization error. The main idea is to divide a space in $\R^{MD}$ to a Cartesian product of $M$ sub-vectors in $\R^D$. The VQ technique is then applied on each group of sub-vectors, resulting in $M$ solutions of K-means in $\R^D$, where each solution has a different set of $K$ clusters. Each vector in $\R^{MD}$ can be encoded using $M \cdot \log_{2}(K)$ bits, by assigning the index of the matching cluster to each of its $M$ sub-vectors. The expressive power of PQ empowers it to transform a vector in $\R^{MD}$ to one of $K^{M}$ possible vectors.

As discussed in Sec.~\ref{sec:inference}, PQ enables an efficient computation of the approximated distance between two compressed vectors using $O(M)$ additions. This is achieved by using $M$ Look Up Tables (LUTs) that store the distance between every two clusters for each of the $M$ partitions. The K-means algorithm is also not bounded by the number of samples, since each of the $\left\{ 1 \dots M \right\}$ k-means solutions partitions space to $K$ clusters, where K is usually small (e.g. $K=256$). Decreasing the quantization error even further, the PQ technique is also able to efficiently compare an uncompressed query vector to a database of compressed vectors. The latter is called asymmetric search, while the former is called symmetric search. Asymmetric search is the common practice in information retrieval systems that employ PQ, since while the database vectors need to be compressed in order to reduce their memory footprint, there is usually no memory limitation for the query, which typically arrives on-the-fly. In PQ, the asymmetric search has been shown to have a lower quantization error, while having the same computational complexity of the symmetric search by constructing LUTs for each query.

The PQ technique has been widely adopted by the information retrieval and computer vision community. It has started a long list of improvements to the original PQ technique. Optimized Product Quantization~\cite{ge2013optimized} (OPQ) and Cartesian K-means~\cite{norouzi2013cartesian} have focused on improving the space decomposition and the learning of the optimal codebooks for decreasing the quantization error. These contributions rely on the observation that simply dividing the features to a Cartesian product does not fully utilize the knowledge about the structure of the feature space, and ignores the intra-subspace correlations of the data. To create a better partition of the space, they suggest to first transform the data by an orthonormal matrix, $R$, and then to do the Cartesian decomposition and learn the optimal clusters. LOPQ~\cite{kalantidis2014locally} used the observation that while PQ and OPQ create an exponential number of possible centroids in $\R^{MD}$, many of them remain without data support, and, therefore, are not used efficiently. To mitigate this problem, they suggest first using a coarse quantizer to cluster the data, and capture its density, and then applying a locally optimized product quantization to each coarse cell.

Despite their tremendous success, Product Quantization techniques and Vector Quantization techniques in general, are being optimized in an unsupervised manner, with the goal of reducing the quantization error. In this work, we further improve Product Quantization techniques by incorporating a supervised signal. Previous works have used supervision to learn a Hamming distance on binary representations, which is a popular alternative technique for ANN.\looseness=-1 

Given two vectors, which are both encoded by $M \cdot \log_{2}(K)$ bits, the possible number of different distance values between them under the Hamming distance is only $M \cdot \log_{2}(K) + 1$. In contrast, the possible number of different distance values between them using PQ is $\binom{K}{2}^M$, which is much larger than Hamming. The richness of the expressive power of PQ has allowed it to outperform Hamming distance techniques that were trained in an unsupervised manner. With the advent of Deep Learning, many binary encoding techniques~\cite{xia2014supervised,lin2015deep,liu2016deep,subic} that utilize end-to-end training and, therefore, benefit from the supervised signal, have been suggested and have proven to be better than the standard PQ technique that is trained in an unsupervised manner~\cite{subic}.

Our work combines the expressive power of the PQ technique with Deep Learning end-to-end optimization techniques, and allows for PQ to benefit from the task-related supervised signal. To our knowledge, we are the first to incorporate a technique inspired by PQ into a deep learning framework. Another work~\cite{cao2016deep} has proposed to combine PQ with Deep Learning for hashing purposes, but in contrast to our work, they do not optimize the clusters of PQ with respect to the supervised signal of classification or retrieval. They alternate instead between learning PQ centroids using K-means on the embeddings space in an unsupervised fashion, and between learning the embedding using a CNN. 
Our solution learns the centroids and the parameters of the CNN end-to-end, while optimizing the centroids explicitly to perform well on classification and retrieval tasks.

While our technique is inspired by Product Quantization, there are a few important technical distinctions. While in PQ the soft representation which is used for asymmetric search is the embedding itself and is not constrained by the vectors of the clusters, in our work as described in Sec.~\ref{sec:dpq}, the soft representation is learned. It is the concatenation of $M$ soft sub-vectors, where each soft sub-vector is a convex combination of the learned centroids. While the asymmetric search capability of PQ improves its performance, it is not explicitly optimized for it, and its success is an outcome of the method's design. In contrast, our method learns both the soft and hard representations as part of the training, and directly improves the asymmetric search. This is done by using a loss function, \emph{joint central loss}, which is inspired by the \emph{center loss}~\cite{wen2016discriminative}. The center loss aims to improve the retrieval performance of a CNN, by learning a center for each class, and adding a term that encourages the embeddings to concentrate around the center of their corresponding class. Our joint central loss is adding another role to center loss, which is to decrease the discrepancy between the soft and the hard representations. This is achieved by optimizing both representations to concentrate around the same class centers.

A structured binary embedding method called SUBIC was recently  proposed~\cite{subic}. In their work, which is the current state of the art for retrieval, each sample is represented by a binary vector of $M K$ bits, where in each group of $K$ bits, only one bit is active. Therefore, each sample can be encoded by $M \cdot \log_{2}(K)$ bits. Similar to other works, the binary representation of SUBIC is not learned explicitly. Instead, each group of $K$ entries is the result of the softmax function, and, therefore, acts as a discrete distribution function on $\{1, \ldots, K\}$. In the inference phase, the entry that corresponds to the highest probability is taken to be the active bit, and all the others are turned into $0$. In order to decrease the discrepancy between the inference and the training, they use regularization to make the distribution function closer to the corners of the simplex (i.e., one-hot vectors). They also enable asymmetric search, by using the original distribution values for the query vector. In contrast, our work learns both the soft and hard representation explicitly, as part of an end-to-end training by using the Straight Through estimator technique~\cite{bengio2013estimating}, and exploits Euclidean distances. This results in a richer expressive power, which improves the classification and retrieval performance, as demonstrated in Sec.~\ref{sec:experiments}.

 \begin{figure}[t]
    \centering
\includegraphics[scale=0.30]{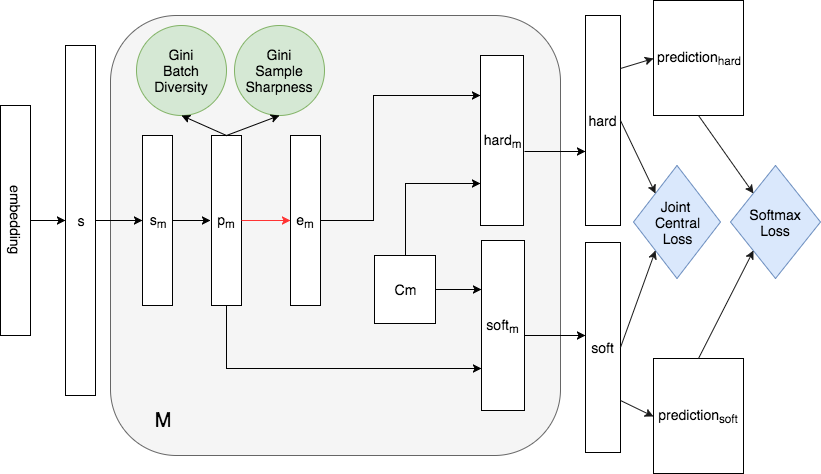}
\caption{\small
\label{fig:DPQ}
\vspace{3pt plus 1pt minus 1pt}
The architecture of the DPQ model. The Softmax Loss and the Joint Central Loss functions are denoted by the blue diamonds, and the Gini Batch Diversity and the Gini Sample Sharpness regularizations are denoted by the green circles. The red arrow is the non-differential one-hot encoding transformation, which requires using the Straight Through estimator, in order to pass the gradients. }
\vspace{-4mm}
\end{figure}

\section{Deep Product Quantization}
\label{sec:dpq}

\paragraph{Architecture.} The diagram of the DPQ architecture is presented in Fig.~\ref{fig:DPQ}. The DPQ is learned on top of the embedding layer. The nature of this embedding changes according to the protocol of each benchmark, see Sec.~\ref{sec:experiments}. Let $x$ be the input to the network, and let $embedding$ be the output of the embedding layer for input $x$ (inputs are omitted for brevity). In the first step, we learn a small multilayer perceptron (MLP) on top of the embedding layer, let $s \in \R^{MN}$ be the output of the MLP. The vector $s$ is then sliced into to $M$ sub-vectors, $s = \left[s_1, s_2, \dots, s_M \right]$, where each $s_m \in \R^N$. On top of each sub-vector, we learn a small MLP which ends in a softmax function with $K$ outputs. We denote the probability of the $k$-th entry of the softmax of the MLP that corresponds to the $m$-th sub-vector by $p_{m}(k)$. 
For each sub-vector, we also learn a matrix, $C_{m} \in \R^{K \times D}$ (composed of $K$ vectors in $\R^D$ that represent the $K$ centroids). We denote the $k$-th row of the matrix $C_{m}$ by $C_{m}(k)$. The $m$-th sub-vector of the soft representation is computed as the convex combination of the rows of $C_{m}$, where the coefficients are the probability values of $p_{m}$:
\begin{equation}
\small
\label{eq:softm}
\soft_m = \sum_{k=1}^{K} {p_{m}(k) \cdot C_{m}(k)}
\end{equation}

Let $k^{*} = \argmax_k p_{m}(k)$ be the index of the highest probability in $p_{m}$, and let $e_m$ be a one hot encoding vector, such that $e_m(k^{*})=1$ and $e_m(k)=0$ for $k \neq k^{*}$. The $m$-th sub-vector of the hard representation is then computed by:
\begin{equation}\label{eq:hardm}
\small
\hard_m = \sum_{k=1}^{K} {e_{m}(k) \cdot C_{m}(k)} = C_{m}(k^{*})
\end{equation}
Therefore, the $m$-th sub-vector of the hard representation is equal to the row in $C_m$ that corresponds to the entry $k^{*}$ with the highest probability in $p_m$. Since the conversion of the probability distribution $p_m$ to a one hot encoding, $e_m$, is not a differential operation, we employ the idea of straight-through (ST) estimator~\cite{bengio2013estimating} to enable the back-propagation, i.e., the computation of the one hot encoding in the forward pass is performed using the $argmax$ function. However, in the backward pass, we treat the one hot encoding layer as the identity function, and pass the gradients received by the one hot encoding layer, directly to the softmax layer that computed $p_m$, without transforming them.

The $M$ soft sub-vectors are concatenated to the final soft representation vector, and the $M$ hard sub-vectors are concatenated to the final hard representation vector:

$\soft = \left[ \soft_1,\dots, \soft_M \right]$,  
$\hard = \left[ \hard_1, \dots, \hard_M \right]$, 
where $\soft$ and $\hard$ are in $\R^{MD}$.\looseness=-1

For classification into $C$ classes, a fully connected layer, defined by a matrix $W \in \R^{MD \times C}$ and a bias vector $b \in \R^C$, is used to obtain prediction scores over these $C$ classes. We denote by  $pred_{\soft}$ and $pred_{\hard}$, the predictions given for the $\soft$ and $\hard$ representations respectively. 
\vspace{-2mm}
\paragraph{Loss functions.} The softmax loss is applied to $pred_{\soft}$ and $pred_{\hard}$ and captures the requirement that the $\soft$ and $\hard$ representations classify the samples correctly. We also devise a new loss function inspired by the center loss~\cite{wen2016discriminative}, named \emph{Joint Central Loss}. 

While the softmax loss encourages the representations to be separable with respect to the classes, the center loss encourages features from the same class to be clustered together, thus improving the discriminative power of the features and contributing to the retrieval performance. The center loss learns a center vector, $o_i \in \R^{V}$, for each class $i$, where $V=MD$ is the size of the representation, by minimizing the distance ${\frac{1}{2}||r_i - o_{y_i}||^2}$ between the representation, $r_i \in R^{V}$, and the vector of the corresponding class, $o_{y_i}$. The motivation for the Joint Central Loss, introduced here, is to add another role to the center loss, which is decreasing the discrepancy between the $\soft$ and $\hard$ representations, thus improving the performance of the asymmetric search. This is implemented by using the same centers for both the $\soft$ and $\hard$ representations,  encouraging both representations to be closer to the same centers of the classes. 
\vspace{-3mm}
\paragraph{Regularization} DPQ uses regularization in order to ensure near uniform distribution of the samples to their corresponding clusters, for each partition $M$. This empowers the training to find a solution that better utilizes the clusters in the encoding. Specifically, given a batch of $B$ samples, $\left( x_1, x_2, ..., x_B \right)$, let $p_m^i \in \R^K$ be the probability distribution over the clusters of the $m$-th sub-vector, of the $i$-th sample. The following Gini Impurity related penalty is then defined as:
\begin{equation}\label{eq:gini}
\small
\mathsf{GiniBatch}(p_m) := \sum_{k=1}^K \left( \frac{1}{B} \sum_{i=1}^B p_{m}^{i}(k) \right)^2
\end{equation}

This penalty achieves a maximal value of $1$ if and only if there is a single cluster, $k$, for which $\forall i\,p_{m}^{i}(k)=1$, and a minimal value of $\frac{1}{K}$ if and only if $\forall k: \frac{1}{B} \sum_{i=1}^{B} p_{m}^{i}(k)=\frac{1}{K}$. 
Therefore, by adding this penalty, the optimization is encouraged to find a solution in which the samples are distributed more evenly to the clusters. 

We also add another regularization term to encourage the probability distribution of a sample $i$, $p_{m}^{i}$, to be closer to a one hot encoding:
\begin{equation}\label{eq:gini2}
\small
\mathsf{GiniSample}(p_m^i) := - \sum_{k=1}^K \left( p_{m}^{i}(k) \right)^2
\end{equation}
This term encourages the soft and hard representations of the same sample to be closer. 
Note that the two loss-functions may seem to be competing. However the first is calculated over a batch and encourages diversity within a batch, while the second is calculated per distribution of a single sample and encourages the distributions to be decisive (i.e., close to a one hot vector).

Similar forms of these regularizations have been successfully used in previous works~\cite{liu2016deep,subic} to improve the performance of hashing techniques.

\subsection{Inference}
\label{sec:inference}
The DPQ method benefits from all the advantages of Product Quantization techniques. This section elaborates on how DPQ is used to create a compressed representation, fast classification, and fast retrieval in both the symmetric and asymmetric forms.

\noindent{\bf Compressed Representation}
For a given vector, $x \in \R^L$, DPQ can compress $x$ to the hard representation. Specifically, $x$ can be encoded by DPQ with $M$ partitions and $K$ clusters per partition, by setting $z = \left( z_1, z_2, \dots, z_M \right)$, where $z_i \in \left\{ 1 \dots K \right\}$ is the cluster to which the i-th partition of $x$ belongs.
Therefore, the hard representation can then be perfectly reconstructed from $z$ and $C_m$, and requires only $M \log_2 (K)$ bits for storage. 
The following compression ratio is achieved when using float-32 to represent $x$: $\frac{32 L}{M \log_2 (K)}$.

\noindent{\bf Classification} By employing Lookup Tables (LUTs), it is possible to decrease the classification time over the hard representation. Let ${pred_{\hard}[c]}$ be the output of the prediction layer for class $c$ according to the hard representation before applying the softmax operation.
\begin{align*}
\small
pred_{\hard}[c] &= b_c + \sum_{d=1}^{MD} {W_{d,c} \cdot \hard[d]}=\\
&= b_c + \sum_{m=1}^{M} \sum_{d=1}^{D} W_{(m-1)D+d,c} \cdot C_{m}(z_m)[d]
\end{align*}

\noindent Using $M$ LUTs of $C \cdot K$ entries, 
$\LUTC_{m}[c,k] = \sum_{d=1}^{D} W_{(m-1)D+d,c} \cdot C_{m}(k)[d],$
one can compute ${pred_{\hard}[c]}$ efficiently by performing $M$ additions:
\begin{equation*}
\small
{pred_{\hard}[c] = b_c + \sum_{m=1}^{M} \LUTC_{m}[c,z_m]}
\end{equation*}

\noindent{\bf Symmetric Comparison} The fast symmetric comparison is performed by using $M$ LUTs, $\LUTSym_{m}[k_1,k_2]$ each of $\binom{K}{2}$ entries:
\begin{equation*}
\small
{\LUTSym_{m}[k_1,k_2]=\sum_{d=1}^{D} \left( C_m(k_1)[d] - C_m(k_2)[d] \right)^2}
\end{equation*}

The distance between the hard representations $\hard^{x}$ and $\hard^{y}$, with compressed hard representations $z^x$ and $z^y$ respectively, can be then computed by:

\begin{equation*}
\small
{\sum_{d=1}^{MD} \left( \hard^{x}[d] - \hard^{y}[d] \right)^2 = \sum_{m=1}^{M} \LUTSym_{m}[z_{m}^{x},z_{m}^{y}]}
\end{equation*}

\noindent{\bf Asymmetric Comparison} The asymmetric comparison is evaluated on the soft representation of a vector, $\soft^{x}$, and on the compressed representation of a vector, $z^y$, that encodes the hard representation of $y$, $\hard^{y}$. The typical use case is when a search system receives a query, computes its soft representation, but uses hard representation to encode the vectors in the database, in order to reduce the memory footprint. In this scenario, it is common to compare the single soft representation of the query with many compressed hard representations of the items in the database. For this application, one can build $M$ LUTs, which are specific to the vector $\soft^{x}$. Each table, $\LUTASym_{m}^{\soft^{x}}$, has $K$ entries:
$\LUTASym_{m}^{\soft^{x}}[k] = \sum_{d=1}^{D} \left( C_m(k)[d] - \soft^{x}[(m-1)\cdot D + d] \right)^2$.
Thus, allowing the comparison of $\soft^{x}$ and $z^y$ by performing $M$ additions:
\begin{equation*}
\small
{\sum_{d=1}^{MD} \left( \soft^{x}[d] - \hard^{y}[d] \right)^2 = \sum_{m=1}^{M} \LUTASym_{m}^{\soft^{x}}[z_{m}^{y}]}
\end{equation*}

\noindent The preprocessing time of preparing the LUT per query, is justified, whenever the database size is much~larger~than~$K$.

\section{Experiments}
\label{sec:experiments}

We evaluate the performance of DPQ on three important tasks: single-domain image retrieval, cross-domain image retrieval, and image classification. Our method is shown to achieve state of the art results in all of them. {\color{black} We employ the same hyper-parameters in each experimental domain, across all the experiments conducted for this domain (size and dataset). As demonstrated in Fig.~\ref{fig:loss}, typically there is a wide range of parameters that produce favorable results. The exact parameters are specified in the supplementary.}
{\color{black}
\subsection{Single-domain category retrieval.} We use the CIFAR-10 dataset to demonstrate the DPQ performance on the single-domain category retrieval task. Since different works have used different evaluation protocols and different features on this dataset, we follow three different protocols that together capture many of the previous works' hashing techniques. We also evaluate DPQ on Imagenet-100 by following the protocol of~\cite{cao2017hashnet,loncaric2018learning}.
\vspace{-2mm}
\paragraph{CIFAR-10 - Protocol 1}
in this protocol, the training set of CIFAR-10 is used for training the model, and the test set is used to evaluate the retrieval performance, employing the mean average precision (mAP) metric. To disentangle the contribution of DPQ from the base architecture of the CNN that is applied on the image, we follow the same architecture proposed by DSH~\cite{liu2016deep}, which was adopted by other works that were evaluated on this benchmark~\cite{liu2016deep,subic}. The protocol of the benchmark is to measure the mAP, when using $12, 24, 36$ and $48$ bits, to encode the database vectors. We train DPQs with $M = 4$ partitions and $K=(8, 64, 512, 4096)$ centroids per partition, to match our experiments with the protocol. DPQ is learned on top of the embedding layer of the base network, that has $U=500$ units. We start by adding a fully connected layer, $F$, on top of $U$, with $V=M \cdot K$ units. We then split  $F\in \R^{V}$ into $M$ equal parts: $F = (F_1, F_2, \dots, F_M)$ where $F_i \in \R^{K}$.
We then apply a softmax function that outputs $p_m$, as described in Sec.~\ref{sec:dpq}, with $K$ entries. Our cluster vectors, $C_m$, are chosen to be in $\R^{Z}$, where Z is a hyper-parameter. In addition to the loss functions and regularizations described in Sec.~\ref{sec:dpq}, we add a weight decay to prevent the over-fitting of the base network. As shown in Tab.~\ref{tab:cifar10}, our DPQ method achieves state of the art results in {\color{black}either} symmetric {\color{black}or} asymmetric retrieval. As mentioned in Sec.~\ref{sec:inference}, both the symmetric and asymmetric methods have the same computation complexity as SUBIC~\cite{subic}. 
Furthermore, we compare our method to the strong and simple baseline suggested in~\cite{sablayrolles2017should}. Since the labels of the database are unknown to the retrieval system, the SSH Classifier+one-hot baseline of~\cite{sablayrolles2017should} is the appropriate baseline to use when evaluating our experiments. In this baseline, one trains a classifier, and uses the binary representation of the class id, as the sample encoding. Therefore, we use the classifier that we trained to encode each sample. Thus, each of the $10$ classes is encoded by $4$ bits. This baseline is indeed very strong as it achieves mAP of $0.627$. However, when training DPQ to use only $4$ bits we are able to surpass the baseline of ~\cite{sablayrolles2017should} by obtaining a mAP of $0.649$. Additionally, this result shows that DPQ with $4$ bits is able to surpass all the other results which are using $12$ bits as shown Tab.~\ref{tab:cifar10}.

\vspace{-6mm}
\paragraph{CIFAR-10 - Protocol 2}
here $10K$ images are selected as queries from the entire $60K$ images of CIFAR-10 ($1K$ from each class). The other $50K$ images are used for training and serve as the database. We follow other methods that were evaluated under this protocol~\cite{li2015feature,zhao2015deep,zhang2015bit,wang2016deep} and use the same architecture and pre-trained weights of VGG-CNN-F~\cite{chatfield2014return} for a fair comparison. We measure the mAP of the algorithm when using $16$, $24$, $32$, and $48$ bits. We train DPQs with $M = (4, 6, 8, 12)$ partitions and $K=16$ centroids per partition, to match our experiments with the protocol. As shown in Tab.~\ref{tab:cifar10_protocol2}, DPQ achieves state of the art results under this protocol.
\vspace{-6mm}
\paragraph{CIFAR-10 - Protocol 3}
The VDSH algorithm~\cite{zhang2016efficient} also uses the architecture and weights of VGG-CNN-F but adopts a different protocol, in which $1000$ images are selected as queries from the $60K$ images of CIFAR10 ($100$ from each class). The other $59K$ images are used for training and serve as the database. We apply DPQ on this protocol and achieve mAP of 0.921 using 16-bits, surpassing the results of VDSH for all the different~bit~settings.
\vspace{-8mm}
\setlength\tabcolsep{3pt}
\begin{table}
\footnotesize	
\centering
\begin{tabular}{|l|c|c|c|c|}
\hline
Method                           & 12-bit & 24-bit & 36-bit & 48-bit \\ \hline \hline
PQ & - & 0.295 & - & 0.290 \\ \hline
PQ-Norm & - & 0.324 & - & 0.319 \\ \hline
LSQ++ (SR-C)~\cite{martinez2018lsq++}     & {-}  & {0.2662} & {-} & {0.2568} \\ \hline
LSQ++ (SR-D)~\cite{martinez2018lsq++}     & {-}  & {0.2578} & {-} & {0.2873} \\ \hline
LSQ++-norm (SR-C)~\cite{martinez2018lsq++}     & {-}  & {0.2868} & {-} & {0.2801} \\ \hline
LSQ++-norm (SR-D)~\cite{martinez2018lsq++}     & {-}  & {0.2662} & {-} & {0.2800} \\ \hline
CNNH+~\cite{xia2014supervised}                             & 0.5425 & 0.5604 & 0.5640 & 0.5574   \\ \hline
DQN~\cite{cao2016deep} & 0.554 & 0.558 & 0.564 & 0.580 \\ \hline
DLBHC~\cite{lin2015deep} & 0.5503 & 0.5803 & 0.5778 & 0.5885  \\ \hline
DNNH~\cite{lai2015simultaneous} & 0.5708 & 0.5875 & 0.5899 & 0.5904          \\ \hline
DSH~\cite{liu2016deep} & 0.6157 & 0.6512 & 0.6607 & 0.675  \\ \hline
KSH-CNN{~\cite{liu2012supervised}} & - & 0.4298 & - & 0.4577 \\ \hline
DSRH~\cite{zhao2015deep} & - & 0.6108 & - & 0.6177 \\ \hline
DRSCH~\cite{zhang2015bit} & - & 0.6219 & - & 0.6305 \\ \hline
BDNN~\cite{do2016learning} & - & 0.6521 & - & 0.6653 \\ \hline
SUBIC~\cite{subic} & 0.6349 & 0.6719 & 0.6823 & 0.6863 \\ \hline
DPQ-Sym     & {\bf 0.7410 }  & {\bf 0.7528 } & {\bf 0.7523 } & {\bf 0.7525} \\ \hline
DPQ-ASym      & {\bf 0.7410 }  & {\bf 0.7543 } & {\bf 0.7539} & {\bf 0.7541} \\ \hline
\end{tabular}
\vspace{1pt plus 1pt minus 1pt}
\caption{Retrieval performance (mAP) on the CIFAR-10 dataset for a varying number of bits. Results for previous methods were copied as is from~\cite{subic}. Missing results that were not reported and are expected not to be competitive, based on the existing ones.}
\label{tab:cifar10}
\vspace{-2mm}
\end{table}

\begin{table}
\footnotesize	
\centering
\begin{tabular}{|l|c|c|c|c|}
\hline
Method                           & 16-bit & 24-bit & 32-bit & 48-bit \\ \hline \hline
DSRH~\cite{zhao2015deep} & 0.608 & 0.611 & 0.617 & 0.618 \\ \hline
DSCH~\cite{zhang2015bit} & 0.609 & 0.613 & 0.617 & 0.62  \\ \hline
DRSCH~\cite{zhang2015bit} & 0.615 & 0.622 & 0.629 & 0.631 \\ \hline
DPSH~\cite{li2015feature} & 0.763 & 0.781 & 0.795 & 0.807 \\ \hline
PQ & 0.846 & 0.849 & 0.849 & 0.851 \\ \hline
PQ-Norm & 0.906 & 0.908 & 0.909 & 0.910 \\ \hline
DTSH~\cite{wang2016deep} & 0.915 & 0.923 & 0.925 & 0.926 \\ \hline
DSDH~\cite{li2017deep} & 0.935 & 0.940 & 0.939 & 0.939 \\ \hline
DPQ-ASym     & { \bf 0.9507 }  & { \bf 0.9508 } & { \bf 0.9507 } & {\bf 0.9507 } \\ \hline

\end{tabular}
\vspace{1pt plus 1pt minus 1pt}
\caption{Retrieval performance (mAP) on CIFAR-10 for a varying bits lengths according to the 2nd protocol. Results for previous methods were copied as is from~\cite{wang2016deep}. }
\label{tab:cifar10_protocol2}
\vspace{-2mm}
\end{table}~\begin{table}
\footnotesize	
\centering
\begin{tabular}{|l|c|c|c|c|}
\hline
Method                           & 16-bit & 32-bit & 64-bit \\ \hline \hline
LSH~\cite{gionis1999similarity} & 0.101 & 0.235 & 0.360  \\ \hline
ITQ~\cite{gong2013iterative} & 0.323 & 0.462 & 0.552  \\ \hline
DHN~\cite{zhu2016deep} & 0.311 & 0.472 & 0.573  \\ \hline
HashNet~\cite{cao2017hashnet} & 0.506 & 0.631 & 0.684  \\ \hline
DBR-v3~\cite{lu2017deep} & 0.733 & 0.761 & 0.769 \\ \hline
HDT~\cite{loncaric2018learning} & 0.838 & 0.822 & 0.812 \\ \hline
DPQ-ASym     & { \bf 0.886 }  & { \bf 0.877 } & { \bf 0.866 }  \\ \hline
\end{tabular}
\vspace{1pt plus 1pt minus 1pt}
\caption{Retrieval performance (mAP@1000) on ImageNet-100 for a varying bits lengths according. Results for previous methods were copied as is from~\cite{loncaric2018learning}. }
\vspace{-6mm}
\label{tab:imagenet-100}

\end{table}

\paragraph{ImageNet-100} 
\vspace{-2mm}
In this protocol, which was suggested by ~\cite{cao2017hashnet}, the training and test sets are drawn from $100$ classes of ImageNet. We follow their experiment and use the same training and test sets definitions. For the base network, we use the same architecture and pre-trained weights of ResNet V2 50~\cite{he2016identity} which ~\cite{loncaric2018learning} has been using. As shown in Tab.\ref{tab:imagenet-100}, DPQ also achieves state of the art results for this dataset.
\begin{table}[t]
\footnotesize	
\centering
\begin{tabular}{|l|c|c|c|}
\hline
Method & VOC2007 & Caltech-101 & ImageNet \\ \hline \hline
PQ~\cite{jegou2011product} & 0.4965 & 0.3089 & 0.1650\\ \hline
CKM~\cite{norouzi2013cartesian} & 0.4995 & 0.3179 & 0.1737\\ \hline
LSQ~\cite{martinez2016revisiting} & 0.4993 & 0.3372 & 0.1882\\ \hline
DSH-64~\cite{liu2016deep} & 0.4914 & 0.2852 & 0.1665\\ \hline
PQ-Norm & 0.5495 & 0.3940 & 0.2229 \\ \hline
LSQ++ (SR-C)~\cite{martinez2018lsq++} & 0.4823 & 0.3735 & 0.1764\\ \hline
LSQ++ (SR-D)~\cite{martinez2018lsq++}  & 0.4824 & 0.3646 & 0.1769\\ \hline
LSQ++-norm (SR-C)~\cite{martinez2018lsq++} & 0.5481 & 0.4122 & 0.2525\\ \hline
LSQ++-norm (SR-D)~\cite{martinez2018lsq++} & 0.5494 & 0.4128 & 0.2534 \\ \hline
SUBIC 2-layer~\cite{subic} & 0.5600 & 0.3923 & 0.2543 \\ \hline
DPQ-Sym 2-layer      & 0.5340  & {{0.4035}} & {{0.3183}} \\ \hline
DPQ-ASym 2-layer      & 0.5371  & {{0.4073}} & {\bf{0.3231}} \\ \hline
DPQ-Sym 2-layer + IN      & {0.5530}  & {{0.4134}} & {{0.3175}} \\ \hline
DPQ-ASym 2-layer + IN      & {\bf{0.5647}}  & {\bf0.4231} & {{0.3227}} \\ \hline
\hline
SUBIC 3-layer~\cite{subic} & 0.5588 & {0.4033} & 0.2810 \\ \hline
DPQ-Sym 3-layer      & 0.5234  & 0.4016 & {{0.3485}} \\ \hline
DPQ-ASym 3-layer      & 0.5292  & {{0.4057}} & {{0.3532}} \\ \hline
DPQ-Sym 3-layer + IN      & {0.5497}  & {{0.4142}} & {{0.3521}} \\ \hline
DPQ-ASym 3-layer + IN      & {\bf 0.5599} & {\bf{0.4253}} & {\bf{0.3557}} \\ \hline
\end{tabular}
\vspace{1pt plus 1pt minus 1pt}
\caption{Retrieval performance (mAP) on the three datasets: ImageNet, Caltech, and VOC2007, where the DPQ model is trained on the ImageNet dataset only, but then evaluated on all three datasets to show cross-domain retrieval. We denote the intra-normalization with IN. }
\label{tab:cross}
\vspace{-4mm}
\end{table}
}
\subsection{Cross-domain category retrieval.}
\label{sub:cross}
In the task of cross-domain category retrieval, one evaluates a supervised hashing technique by training on a dataset with specific classes, and evaluating the retrieval results by using the mAP metric on a different dataset with a different set of classes. The authors of~\cite{sablayrolles2017should} have demonstrated the importance of using this task for the evaluation of a hashing technique, in addition to the standard single-domain category retrieval. 
\vspace{-2mm}
\paragraph{Protocol} We follow the protocol of SUBIC~\cite{subic}, and train a DPQ model on vectors in $\R^{128}$, which were computed by applying the VGG-128~\cite{chatfield2014return} pre-trained model on the ILSVRC-ImageNet dataset and extracting the embedding representation. The DPQ model applies a fully connected layer with $2048$ units on the input, and then applies the ReLU activation. We then split the resulting vector to eight equal sub-vectors, each in $\R^{256}$. For each sub-vector, we apply the softmax function which outputs $p_m$, as described in Sec.~\ref{sec:dpq}, with $K=256$ entries. Therefore, our DPQ encodes each vector into $64$ bits in the compressed hard representation. Our cluster vectors, $C_m$, are chosen to be in $\R^{64}$. {\color{black} In~\cite{subic} two feature types were used: 2-layer and 3-layer. The 2-layer experiments are trained on the embedding representation of VGG-128~\cite{chatfield2014return}, and the 3-layer experiments are trained on the representation from the layer, prior to the embedding layer.}
\begin{figure}[t]
  \small
    \centering
\includegraphics[width=8.2cm,height=5cm]{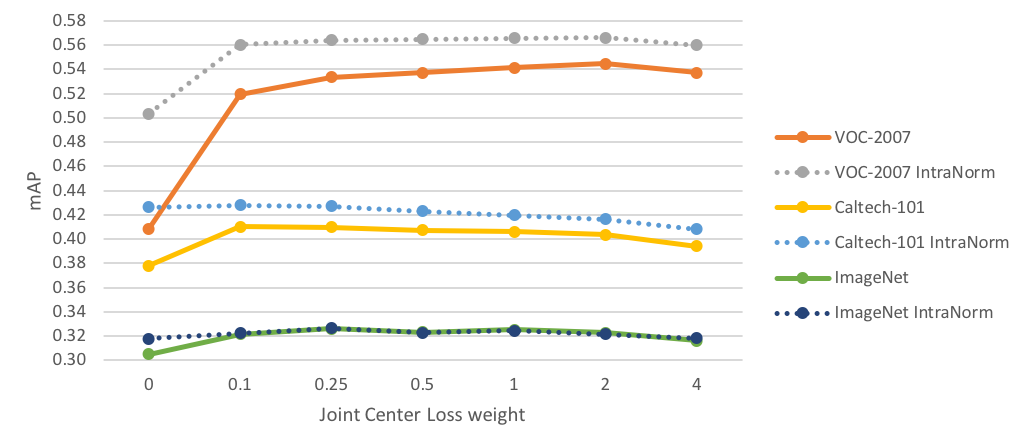}
fF\caption{\small
\label{fig:loss}
The retrieval performance (mAP) for the cross-domain category retrieval benchmark as a function of the Joint Central Loss weight. The DPQ model is trained on the ImageNet dataset, and is evaluated on three different datasets: VOC2007, Caltech-101, and ImageNet. As shown, The Joint Central Loss is improving the results on all the different datasets. Furthermore, the intra-normalization is improving the results for the cross-domain datasets of VOC2007 and Caltech-101, while not affecting the performance of ImageNet.
The reported results are for the 2-layer asymmetric case.
}
\vspace{-4mm}
\end{figure}
We then evaluate the performance of hashing using DPQ for retrieval on the ImageNet validation set, and on the Caltech-101 and VOC2007 datasets. Following~\cite{subic}, we use $1000$, $1000$ and $2000$ random query images from the datasets of Caltech-101, VOC2007, and ImageNet respectively, and use the rest as the database. Our results are presented in Tab.~\ref{tab:cross}. Our method surpasses the state of the art result, {\color{black}for both the 2-layer and the 3-layer cases,} for the ImageNet {\color{black} and Caltech-101 datasets, but as-is not on VOC2007.
{\color{black}To further support cross-domain hashing,} we developed an intra-normalization technique for our soft and hard representations, which was inspired by the intra-normalization technique of~\cite{arandjelovic2013all}. That method improves the retrieval obtained with a VLAD based representation, which was trained on top of SIFT features of one dataset, but then applied to another. Specifically, we perform $L2$ normalization for each $\hard_m$ and for each $\soft_m$, resulting in ${\hardnorm_m}$ and ${\softnorm_m}$ respectively. We     then concatenate them and produce the new $\hard$ and $\soft$ representations. Note that performing the $L2$ normalization to each sub-vector $m = 1 \dots M$ separately, instead of performing $L2$ normalization to the entire $\hard$ and $\soft$ representations, does not hurt our ability to use LUTs for inference, as described in Sec.~\ref{sec:inference}. One can simply replace the clusters of $C_m$ with their normalized version. 

The intra-normalization almost does not affect the ImageNet evaluation,  which is a single-domain category retrieval task. 
As shown in Tab.~\ref{tab:cross}, the asymmetric search outperforms the symmetric search. Together with the intra-normal\-ization technique, we {\color{black} improve our results on 
both VOC2007 and Caltech-101}. Similar to SUBIC~\cite{subic}, the 3-layer experiments show substantial improvement on the ImageNet dataset, with respect to the 2-layer experiments.

As a baseline, we conducted another experiment in which the $L2$ normalization is performed as part of the model training using an $L2$ normalization layer. This experiment resulted in inferior results on the Caltech-101 and VOC2007 datasets, with respect to training without $L2$ normalization and applying the intra-normalization technique.

In order to study the importance of the joint central loss, we depict in Fig.~\ref{fig:loss} the mAP for the cross domain category retrieval benchmark  as a function of the weight assigned to this loss. As can be seen, when training DPQ with a joint central loss of weight 0.1, a significant increase in mAP is observed across datasets. The mAP very gradually decreases, as this weight further increases. 
\vspace{-2mm}
{\color{black}
\vspace{4mm}
\paragraph{A simple unsupervised strong baseline}
\vspace{-2mm}
A simple but strong unsupervised baseline that we discovered is performing product quantization of the normalized features of VGG, instead of the original features. To be more precise, we normalize the features be on the unit sphere before performing product quantization. This is, in some sense, equivalent to having the product quantization estimate the cosine distance between the features, instead of the euclidean distance. As shown in Tab.~\ref{tab:cross} this simple unsupervised baseline that we denote as PQ-Norm, achieves substantial improvement over the product quantization that was trained on the original features, and performs slightly worse than supervised methods, such as SUBIC~\cite{subic} and ours.
}
\setlength\tabcolsep{2pt}
\begin{table}
\footnotesize	
\centering
\begin{tabular}{|l|c|c|}
\hline
& \multicolumn{2}{c|}{ImageNet} \\ \hline 
Method & Top-1 Accuracy & Top-5 Accuracy \\ \hline 
PQ~\cite{jegou2011product} & 39.88 & 67.22 \\ \hline
CKM~\cite{norouzi2013cartesian}  & 41.15 & 69.66 \\ \hline
SUBIC~\cite{subic} & 47.77 & 72.16 \\ \hline
DPQ & {\bf{56.80}} & {\bf{77.59}} \\ \hline
\end{tabular}
\vspace{1pt plus 1pt minus 1pt}
\caption{Classification performance on ImageNet using learned 64-bit representations}
\label{tab:hashclassification}
\vspace{-2mm}
\end{table}

\subsection{Image classification} As discussed in Sec.~\ref{sec:inference}, 
DPQ can efficiently classify samples given their compressed representation.
We follow the protocol of SUBIC~\cite{subic}, and report the Top-1 and Top-5 accuracy on the test set of ImageNet, using the 64-bit compressed hard representation. As depicted in Tab.~\ref{tab:hashclassification}, our DPQ method surpasses the state of the art.
\begin{table}
\footnotesize	
\setlength\tabcolsep{2pt}
\centering
\begin{tabular}{ | c | c | c | }
\hline
Method & Oxford5K & Paris6K \\ \hline
PQ~\cite{jegou2011product} & 0.2374 & 0.3597 \\ \hline
LSQ~\cite{martinez2016revisiting} & 0.2512 & 0.3764 \\ \hline
DSH-64~\cite{liu2016deep} & 0.2108 & 0.3287 \\ \hline
SUBIC~\cite{subic} & 0.2626 & 0.4116 \\ \hline
DPQ (ours) & \bf{0.2643} & \bf{0.4249} \\ \hline
PQ-Norm & \bf{0.2646 $\pm$ 0.0012} & \bf{0.4262 $\pm$ 0.0036}  \\ \hline
\end{tabular}
\newline
\caption{Retrieval performance (mAP) on the Oxford5K and Paris6K datasets, according to the protocol defined in~\cite{subic}. The first four lines were copied as is from~\cite{subic}. The results in the last line were calculated by running PQ $5$ times using $5$ random seeds for both datasets. The mean and standard deviation are reported.}
\label{tab:instret}
\vspace{-4mm}
\end{table}

{\color{black}
\subsection{Instance retrieval based on landmarks} SUBIC~\cite{subic} has reported an improvement in retrieval over PQ on the Oxford~\cite{philbin2007object} and the Paris~\cite{philbin2008lost} benchmarks, when training on the clean train~\cite{gordo2016deep} subset of the landmarks dataset~\cite{babenko2014neural}, using the features extracted from the embedding layer of VGG-128. Our system obtains slightly better results than SUBIC on these benchmarks. However, rerunning the baselines, with FAISS~\cite{JDH17} implementation of the PQ method on the normalized features that were used by SUBIC~\cite{subic}, results with performance which is on par with our method as shown in Tab.~\ref{tab:instret}. This further supports the simple baseline that was introduced in Sec.~\ref{sub:cross}.
\vspace{-2mm}
\subsection{All pairwise distances} 
\label{subsec:all}
The performance of DPQ for both the symmetric and asymmetric retrieval is very similar. An application for which the quality of the symmetric retrieval is highly important, is the \emph{all pairwise distances}. In this application we want to compute the distance between every two samples in the database. Since all of the items in the database are compressed, the asymmetric version is not available and, therefore, one must rely on the quality of the symmetric search.  The expressive power of DPQ defines $\binom{K}{2}^M$ possible distances between two hard representations of vectors. In contrast, the Hamming distance on $M \cdot \log_{2}(K)$ bits, defines $M \cdot \log_{2}(K) + 1$ possible distances between two binary vectors. In SUBIC~\cite{subic}, the hard representation is structured such that each group, $m \in\{ 1, \ldots , M\}$, has only one bit that is active, therefore allowing only $M+1$ possible values of distances between two hard representations. To validate our hypothesis, we evaluate SUBIC~\cite{subic} on the VOC2007 dataset using the code provided by SUBIC and measured the mAP when using the symmetric search. This resulted in a mAP of 0.4443, which is lower than their asymmetric search result of 0.56 and lower than the unsupervised techniques, as shown in Tab.~\ref{tab:cross}. However, our symmetric retrieval achieves a mAP of 0.5530 and does not fall far from our asymmetric retrieval performance.
\vspace{-3mm}
\section{Conclusion}
Our approach is supervised and extends the unsupervised Product Quantization technique by building LUTs, which are learned from the features and the labels. Our method is directly optimized for the retrieval of the asymmetric search, since it learns both the soft and hard representations as part of the training. Furthermore, as shown in Sec.~\ref{sec:experiments}, the symmetric search performance of DPQ does not fall too far behind the asymmetric search performance. This has an advantage, for example, in cases where one is interested in performing all versus all comparisons on a compressed database. This is contrast to some methods, such as ~\cite{subic} which has a large gap between its asymmetric and symmetric performance, as shown in Sec.~\ref{subsec:all}.

While having the same memory footprint and inference time as Product Quantization, our experiments show that DPQ achieves state of the art results in multiple benchmarks that are commonly used in the literature.

\section*{Acknowledgements}
This project has received funding from the European Research Council (ERC) under the European Unions Horizon 2020 research and innovation programme (grant ERC CoG 725974). The contribution of the first author is part of a Ph.D. thesis research conducted at Tel Aviv University.  

}
{\small
\bibliographystyle{ieee}
\bibliography{cameraready}

\begin{thebibliography}{10}\itemsep=-1pt

\bibitem{arandjelovic2013all}
R.~Arandjelovic and A.~Zisserman.
\newblock All about vlad.
\newblock In {\em Proceedings of the IEEE conference on Computer Vision and
  Pattern Recognition}, pages 1578--1585, 2013.

\bibitem{babenko2014neural}
A.~Babenko, A.~Slesarev, A.~Chigorin, and V.~Lempitsky.
\newblock Neural codes for image retrieval.
\newblock In {\em {ECCV}}, pages 584--599. Springer, 2014.

\bibitem{bengio2013estimating}
Y.~Bengio, N.~L{\'e}onard, and A.~Courville.
\newblock Estimating or propagating gradients through stochastic neurons for
  conditional computation.
\newblock {\em arXiv preprint arXiv:1308.3432}, 2013.

\bibitem{cao2016deep}
Y.~Cao, M.~Long, J.~Wang, H.~Zhu, and Q.~Wen.
\newblock Deep quantization network for efficient image retrieval.
\newblock In {\em AAAI}, pages 3457--3463, 2016.

\bibitem{cao2017hashnet}
Z.~Cao, M.~Long, J.~Wang, and S.~Y. Philip.
\newblock Hashnet: Deep learning to hash by continuation.
\newblock In {\em ICCV}, pages 5609--5618, 2017.

\bibitem{chatfield2014return}
K.~Chatfield, K.~Simonyan, A.~Vedaldi, and A.~Zisserman.
\newblock Return of the devil in the details: Delving deep into convolutional
  nets.
\newblock {\em arXiv preprint arXiv:1405.3531}, 2014.

\bibitem{do2016learning}
T.-T. Do, A.-D. Doan, and N.-M. Cheung.
\newblock Learning to hash with binary deep neural network.
\newblock In {\em {ECCV}}, pages 219--234. Springer, 2016.

\bibitem{ge2013optimized}
T.~Ge, K.~He, Q.~Ke, and J.~Sun.
\newblock Optimized product quantization for approximate nearest neighbor
  search.
\newblock In {\em {CVPR}}, pages 2946--2953, 2013.

\bibitem{gionis1999similarity}
A.~Gionis, P.~Indyk, R.~Motwani, et~al.
\newblock Similarity search in high dimensions via hashing.
\newblock In {\em Vldb}, volume~99, pages 518--529, 1999.

\bibitem{gong2013iterative}
Y.~Gong, S.~Lazebnik, A.~Gordo, and F.~Perronnin.
\newblock Iterative quantization: A procrustean approach to learning binary
  codes for large-scale image retrieval.
\newblock {\em IEEE Transactions on Pattern Analysis and Machine Intelligence},
  35(12):2916--2929, 2013.

\bibitem{gordo2016deep}
A.~Gordo, J.~Almaz{\'a}n, J.~Revaud, and D.~Larlus.
\newblock Deep image retrieval: Learning global representations for image
  search.
\newblock In {\em {ECCV}}, pages 241--257. Springer, 2016.

\bibitem{gray1998quantization}
R.~M. Gray and D.~L. Neuhoff.
\newblock Quantization.
\newblock {\em IEEE transactions on information theory}, 44(6):2325--2383,
  1998.

\bibitem{he2016identity}
K.~He, X.~Zhang, S.~Ren, and J.~Sun.
\newblock Identity mappings in deep residual networks.
\newblock In {\em European conference on computer vision}, pages 630--645.
  Springer, 2016.

\bibitem{subic}
H.~Jain, J.~Zepeda, P.~Perez, and R.~Gribonval.
\newblock Subic: A supervised, structured binary code for image search.
\newblock In {\em {ICCV}}.

\bibitem{jegou2011product}
H.~Jegou, M.~Douze, and C.~Schmid.
\newblock Product quantization for nearest neighbor search.
\newblock {\em IEEE transactions on pattern analysis and machine intelligence},
  33(1):117--128, 2011.

\bibitem{JDH17}
J.~Johnson, M.~Douze, and H.~J{\'e}gou.
\newblock Billion-scale similarity search with gpus.
\newblock {\em arXiv preprint arXiv:1702.08734}, 2017.

\bibitem{kalantidis2014locally}
Y.~Kalantidis and Y.~Avrithis.
\newblock Locally optimized product quantization for approximate nearest
  neighbor search.
\newblock In {\em {CVPR}}, pages 2321--2328, 2014.

\bibitem{lai2015simultaneous}
H.~Lai, Y.~Pan, Y.~Liu, and S.~Yan.
\newblock Simultaneous feature learning and hash coding with deep neural
  networks.
\newblock In {\em CVPR}, pages 3270--3278, 2015.

\bibitem{li2017deep}
Q.~Li, Z.~Sun, R.~He, and T.~Tan.
\newblock Deep supervised discrete hashing.
\newblock In {\em Advances in Neural Information Processing Systems}, pages
  2482--2491, 2017.

\bibitem{li2015feature}
W.-J. Li, S.~Wang, and W.-C. Kang.
\newblock Feature learning based deep supervised hashing with pairwise labels.
\newblock {\em arXiv preprint arXiv:1511.03855}, 2015.

\bibitem{liu2016deep}
H.~Liu, R.~Wang, S.~Shan, and X.~Chen.
\newblock Deep supervised hashing for fast image retrieval.
\newblock In {\em CVPR}, pages 2064--2072, 2016.

\bibitem{liu2012supervised}
W.~Liu, J.~Wang, R.~Ji, Y.-G. Jiang, and S.-F. Chang.
\newblock Supervised hashing with kernels.
\newblock In {\em Computer Vision and Pattern Recognition (CVPR), 2012 IEEE
  Conference on}, pages 2074--2081. IEEE, 2012.

\bibitem{loncaric2018learning}
M.~Loncaric, B.~Liu, and R.~Weber.
\newblock Learning hash codes via hamming distance targets.
\newblock {\em arXiv preprint arXiv:1810.01008}, 2018.

\bibitem{lu2017deep}
X.~Lu, L.~Song, R.~Xie, X.~Yang, and W.~Zhang.
\newblock Deep binary representation for efficient image retrieval.
\newblock {\em Advances in Multimedia}, 2017, 2017.

\bibitem{martinez2016revisiting}
J.~Martinez, J.~Clement, H.~H. Hoos, and J.~J. Little.
\newblock Revisiting additive quantization.
\newblock In {\em ECCV}, pages 137--153. Springer, 2016.

\bibitem{martinez2018lsq++}
J.~Martinez et~al.
\newblock {LSQ++}: Lower running time and higher recall in multi-codebook
  quantization.
\newblock In {\em ECCV}, 2018.

\bibitem{norouzi2013cartesian}
M.~Norouzi and D.~J. Fleet.
\newblock Cartesian k-means.
\newblock In {\em CVPR}, pages 3017--3024, 2013.

\bibitem{philbin2007object}
J.~Philbin, O.~Chum, M.~Isard, J.~Sivic, and A.~Zisserman.
\newblock Object retrieval with large vocabularies and fast spatial matching.
\newblock In {\em CVPR}, pages 1--8. IEEE, 2007.

\bibitem{philbin2008lost}
J.~Philbin, O.~Chum, M.~Isard, J.~Sivic, and A.~Zisserman.
\newblock Lost in quantization: Improving particular object retrieval in large
  scale image databases.
\newblock In {\em CVPR}, pages 1--8. IEEE, 2008.

\bibitem{sablayrolles2017should}
A.~Sablayrolles, M.~Douze, N.~Usunier, and H.~J{\'e}gou.
\newblock How should we evaluate supervised hashing?
\newblock In {\em Acoustics, Speech and Signal Processing (ICASSP), 2017 IEEE
  International Conference on}, pages 1732--1736. IEEE, 2017.

\bibitem{wang2016deep}
X.~Wang, Y.~Shi, and K.~M. Kitani.
\newblock Deep supervised hashing with triplet labels.
\newblock In {\em Asian Conference on Computer Vision}, pages 70--84. Springer,
  2016.

\bibitem{wen2016discriminative}
Y.~Wen, K.~Zhang, Z.~Li, and Y.~Qiao.
\newblock A discriminative feature learning approach for deep face recognition.
\newblock In {\em European Conference on Computer Vision}, pages 499--515.
  Springer, 2016.

\bibitem{xia2014supervised}
R.~Xia, Y.~Pan, H.~Lai, C.~Liu, and S.~Yan.
\newblock Supervised hashing for image retrieval via image representation
  learning.
\newblock In {\em AAAI}, volume~1, pages 2156--2162, 2014.

\bibitem{zhang2015bit}
R.~Zhang, L.~Lin, R.~Zhang, W.~Zuo, and L.~Zhang.
\newblock Bit-scalable deep hashing with regularized similarity learning for
  image retrieval and person re-identification.
\newblock {\em IEEE Transactions on Image Processing}, 24(12):4766--4779, 2015.

\bibitem{zhang2016efficient}
Z.~Zhang, Y.~Chen, and V.~Saligrama.
\newblock Efficient training of very deep neural networks for supervised
  hashing.
\newblock In {\em Proceedings of the IEEE Conference on Computer Vision and
  Pattern Recognition}, pages 1487--1495, 2016.

\bibitem{zhao2015deep}
F.~Zhao, Y.~Huang, L.~Wang, and T.~Tan.
\newblock Deep semantic ranking based hashing for multi-label image retrieval.
\newblock In {\em Computer Vision and Pattern Recognition (CVPR), 2015 IEEE
  Conference on}, pages 1556--1564. IEEE, 2015.

\bibitem{zhu2016deep}
H.~Zhu, M.~Long, J.~Wang, and Y.~Cao.
\newblock Deep hashing network for efficient similarity retrieval.
\newblock In {\em AAAI}, pages 2415--2421, 2016.

\end{thebibliography}
}

\end{document}